\documentclass{article}
\usepackage{ijcai22}

\usepackage{times}
\usepackage{soul}
\usepackage{url}
\usepackage[hidelinks]{hyperref}
\usepackage[utf8]{inputenc}
\usepackage[small]{caption}
\usepackage{graphicx}
\usepackage{amsmath}
\usepackage{amsthm}
\usepackage{amsfonts}
\usepackage{booktabs}
\usepackage{algorithm}
\usepackage{algorithmic}
\usepackage{bm}

\usepackage{dsfont}

\usepackage{xcolor}
\usepackage{subcaption}

\newtheorem{definition}{Definition}

\title{On Attacking Out-Domain Uncertainty Estimation in Deep Neural Networks}

\author{
Huimin Zeng$^1$\footnote{Contact Author}\and
Zhenrui Yue$^1$\and
Yang Zhang$^2$\and
Ziyi Kou$^1$\and
Lanyu Shang$^1$\and
Dong Wang$^1$\\
\affiliations
$^1$Unversity of Illinois at Urbana-Champaign\\
$^2$University of Notre Dame\\
\emails
\{huiminz3, zhenrui3, ziyikou2, lshang3, dwang24\}@illinois.edu,
yzhang42@nd.edu
}

\begin{document}

\maketitle

\begin{abstract}
In many applications with real-world consequences, it is crucial to develop reliable uncertainty estimation for the predictions made by the AI decision systems. Targeting at the goal of estimating uncertainty, various deep neural network (DNN) based uncertainty estimation algorithms have been proposed. However, the robustness of the uncertainty returned by these algorithms has not been systematically explored. In this work, to raise the awareness of the research community on robust uncertainty estimation, we show that state-of-the-art uncertainty estimation algorithms could fail catastrophically under our proposed adversarial attack despite their impressive performance on uncertainty estimation. In particular, we aim at attacking the out-domain uncertainty estimation: under our attack, the uncertainty model would be fooled to make high-confident predictions for the out-domain data, which they originally would have rejected. Extensive experimental results on various benchmark image datasets show that the uncertainty estimated by state-of-the-art methods could be easily corrupted by our attack.
\end{abstract}

\section{Introduction}
\label{sec:intro}
Deep neural networks (DNNs) have been thriving in various applications, such as computer vision, natural language processing and decision making. However, in many applications with real-world consequences (e.g. autonomous driving, disease diagnosis, loan granting), it is not sufficient to merely pursue the high accuracy of the predictions made by the AI models, since the deterministic wrong predictions without any uncertainty justification could lead to catastrophic consequences \cite{galil2021disrupting}. Therefore, to address the issue of producing over-confident wrong predictions from DNNs, great efforts have been made to quantify the predictive uncertainty of the models, so that ambiguous or low-confident predictions could be rejected or deferred to an expert. 

Indeed, state-of-the-art algorithms for uncertainty estimation in DNNs have shown their impressive performance in terms of quantifying the confidence of the model predictions. Handling either in-domain data (generated from the training distribution) or out-domain data under domain shift, these algorithms could successfully assign low confidence scores for ambiguous predictions. However, the robustness of such estimated uncertainty/confidence is barely studied. The robustness of uncertainty estimation in this paper is defined as: to which extent would the predictive confidence be affected when the input is deliberately perturbed? Consider an example of autonomous driving \cite{feng2018towards}, where the visual system of a self-driving car is trained using collected road scene images. When an extreme weather occurs, to avoid over-confident wrong decisions, the visual system would show high uncertainty for the images captured by the sensors, since the road scenes observed by the sensors under the extreme weather (out-domain) could be drastically different from the training scenes (in-domain). However, a malicious attacker might attempt to perturb the images captured by the sensors in such a way, that the visual system would regard the out-domain images as in-domain images, and make completely wrong decisions with a high confidence, leading to catastrophic consequences.


\begin{figure*}[ht]
\centering
\begin{subfigure}{0.85\textwidth}
  \includegraphics[width=\linewidth]{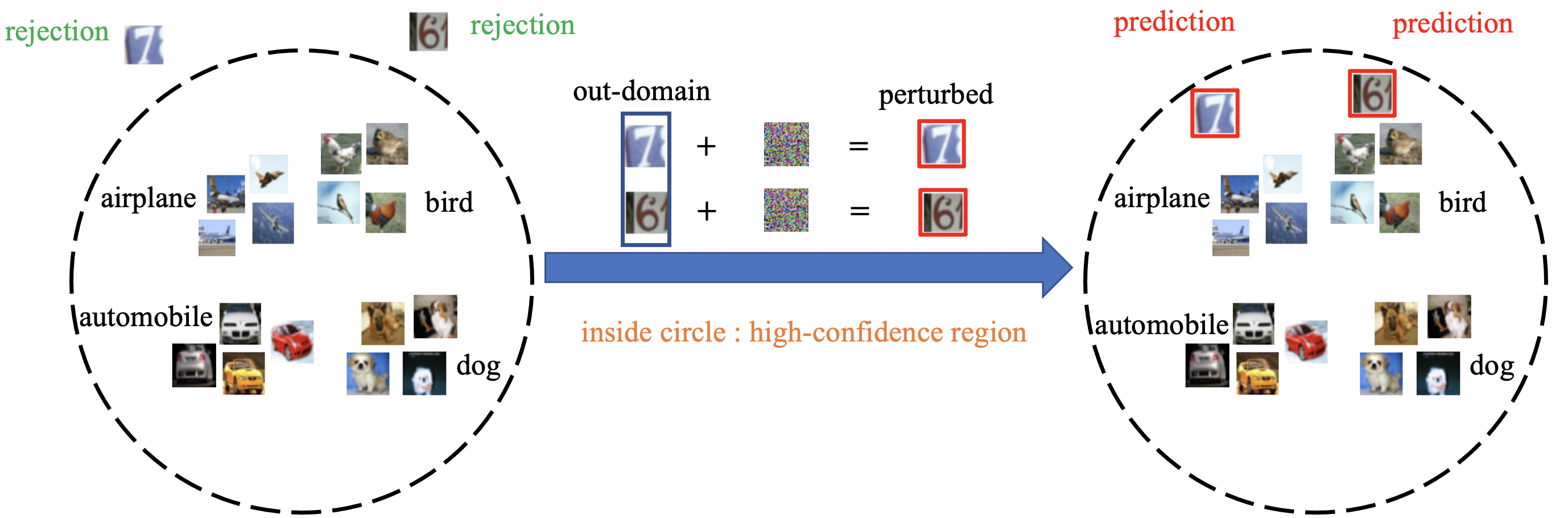}
\end{subfigure}
\caption{Consider an uncertain model trained on images of airplanes, automobile, birds and dogs. }
\label{fig:intuition}
\end{figure*}

Therefore, with the intention of raising the attention of the research community to systematically investigate the \textit{robustness of the uncertainty estimation}, we show that SoTA DNN-based uncertainty estimation algorithms could fail easily under our proposed adversarial attack. In particular, we focus on attacking the uncertainty estimation for \emph{out-domain} data in an image classification problem. That is, under our attack, the uncertainty estimation models would be fooled to assign extremely high confidence scores for the out-domain images, which originally would have been rejected by these models due to the low confidence scores (e.g. Softmax score \cite{galil2021disrupting}). As shown in Figure~\ref{fig:intuition}, the attacker will perturb the out-domain images into the high-confidence region of the victim system. To show vulnerability of the SoTA DNN based uncertainty estimation algorithms under our threat model, we launched our proposed out-domain adversarial attack on various algorithms, including Deep Ensemble \cite{lakshminarayanan2016simple}, RBF-based Deterministic Uncertainty Quantification (DUQ) \cite{van2020uncertainty}, Gaussian process based Deterministic Uncertainty Estimation (DUE) \cite{van2021feature} and Spectral-Normalized Gaussian Process (SNGP) \cite{liu2020simple}. Extensive experimental results in various benchmark image datasets show that the uncertainty estimated by these algorithms could be drastically corrupted under our attack.


\section{Related Work}
\label{sec:rela}
\paragraph{Uncertainty Estimation in DNNs.} A significant amount of algorithms have been proposed to quantify the uncertainty in DNNs \cite{lakshminarayanan2016simple,van2021feature,van2020uncertainty,liu2020simple,alarab2021adversarial,gal2016dropout}. For instance, in \cite{lakshminarayanan2016simple}, a set of neural networks were trained to construct an ensemble model for uncertainty estimation using the training data. In \cite{van2020uncertainty}, RBF (radial basis function) kernel was used to quantify the uncertainty expressed by the deep neural networks. In addition, a further thread of studies built their algorithm based on the Gaussian process (GP), which has been proved to a powerful tool for uncertainty estimation theoretically. However, GP suffers from low expressive power, resulting in poor model accuracy \cite{bradshaw2017adversarial}. Therefore, based on \cite{bradshaw2017adversarial}, \cite{van2021feature} proposed to regularize the feature extractor, so that the deep kernel learning could be stabilized. Similarly, in \cite{liu2020simple}, the predictive uncertainty is computed with a spectral-normalized feature extractor and a Gaussian process. However, the \emph{robustness} of these state-of-the-art algorithms is not well studied. As shown in this paper, the uncertain DNN models trained using these algorithms could be fooled easily under our attack.

\paragraph{Adversarial Examples.} Despite the impressive performance on clean data, it has been shown that DNNs could be extremely vulnerable to adversarial examples \cite{szegedy2013intriguing,goodfellow2014explaining,kannan2018adversarial,carlini2017adversarial}. That is, deep classifiers could be fooled to make completely wrong predictions for the input images that are deliberately perturbed. However, our proposed adversarial attack is different from the traditional adversarial examples in two aspects. Firstly, traditional adversarial examples are crafted by the attacker to corrupt the accuracy of classifiers, whereas our threat model is aimed at attacking uncertainty estimation in DNNs and increasing its confidence of the wrong predictions. Moreover, the traditional adversarial attack is defined under the close-world assumption \cite{reiter1981closed,han2021iterative}, where all possible predictive categories are covered by the training data. The in-domain property limits the power of adversary, since all possible perturbing directions could be simulated by performing untargeted adversarial training \cite{carlini2017adversarial}. In comparison, our proposed attack is out-domain: perturbing the out-domain data into the in-domain data distribution that the uncertainty estimation model is trained to fit. As shown in our experiments, traditional in-domain adversarial defense mechanism could not withstand our attack: even adversarially trained models could be fooled by our out-domain adversarial examples.


\paragraph{Adversarial Attacks on Uncertainty Estimation.} The idea of attacking uncertainty estimation in DNNs have been introduced in for the first time \cite{galil2021disrupting}. In \cite{galil2021disrupting}, the authors tried to manipulate the confidence score over the predictions by perturbing the input images. By heuristically moving the correctly classified input images towards the decision boundary of the model, the attacker could reduce the confidence of the model on the correct predictions. However, the threat model in \cite{galil2021disrupting} could only attack in-domain images, whereas our attack is formulated for an out-domain scenario. Moreover, the adversarial robustness of OOD models has been systematically investigated in \cite{meinke2021provably,augustin2020adversarial,meinke2019towards,hein2019relu,bitterwolf2020certifiably}. However, compared to \cite{meinke2021provably,augustin2020adversarial,meinke2019towards,hein2019relu,bitterwolf2020certifiably}, where ReLU-based classifiers are mainly discussed, this work aims at designing a more general threat model for broader range of uncertainty estimation algorithms.

\section{Problem Statement}
\label{sec:pro}
\label{sec:section}
\subsection{Key Concept Definition}

\begin{definition}[Data Domain]
We define two data domains in this paper, namely $\mathcal{P}_{\mathrm{IN}}$ (representing in-domain distribution) and $\mathcal{P}_{\mathrm{OUT}}$ (for out-domain distribution).
\end{definition}
An excellent uncertainty estimation model is expected to make high-confident predictions for the in-domain data, but produce low-confident predictions for out-domain data.

\begin{definition}[Dataset]
In our problem, the training set $\mathcal{D}_{train}$ consists of images sampled from $\mathcal{P}_{\mathrm{IN}}$: $\mathcal{D}_{train} = \{(\bm{x},y)|(\bm{x},y) \sim \mathcal{P}_{\mathrm{IN}} \}$, where $\bm{x}$ refers to input images and $y$ represents the ground truth labels for $\bm{x}$. The test set contains images sampled from $\mathcal{P}_{\mathrm{OUT}}$, since we mainly evaluate our attack on out-domain data: $\mathcal{D}_{test} = \{\bm{x}|\bm{x} \sim \mathcal{P}_{\mathrm{OUT}} \}$. 
\end{definition}

\begin{definition}[Uncertainty Estimation] Most uncertainty estimation models $f_{\bm{\theta}}$ are defined as a mapping from the input image $\bm{x}$ to the predictive confidence distributed over all possible categories $\bm{\hat{p}(y)}$: $f_{\bm{\theta}}: \bm{x} \to \bm{\hat{p}(y)} \in \bm{\mathcal{R}}^{C}$, where $C$ is the total number of classes of the in-domain data. 
\end{definition}
Regarding $\bm{\hat{p}(y)}$, an ideal uncertainty estimation model will assign a uniform confidence over all classes for the out-domain data, since it would be too uncertain to make any predictions but random guess. Moreover, dependent on uncertainty estimation algorithm, $\bm{\hat{p}(y)}$ could have different physical implications. For instance, in Deep Ensemble, $\bm{\hat{p}(y)}$ is the Softmax uncertainty score. In DUQ, $\bm{\hat{p}(y)}$ is interpreted as the distances between the test data point and the centroids.

\begin{definition}[Evaluation Metrics] In terms of evaluating the model's uncertainty on out-domain data, we use two metrics:
\item Entropy $\mathcal{H}(f_{\bm{\theta}})$:
\vspace{-0.2cm}
\begin{equation}
\begin{split}
\label{eq:entropy}
    \mathcal{H}(f_{\bm{\theta}}) &= \mathbb{E}_{\bm{x}\sim \mathcal{\bm{D}}_{test}} \Big[ \sum_{i=1}^{C} - [f_{\bm{\theta}}(\bm{x})]_i log [f_{\bm{\theta}}(\bm{x})]_i \Big ]\\
    &= \mathbb{E}_{\bm{x}\sim \mathcal{\bm{D}}_{test}} \Big[ \sum_{i=1}^{C} -\hat{p}(y_i) log \hat{p}(y_i) \Big].
\end{split}
\end{equation}


\item Rejection Rate $\mathcal{R}(f_{\bm{\theta}})$:
\vspace{-0.2cm}
\begin{equation}
\begin{split}
\label{eq:rejection_rate}
    \mathcal{R}(f_{\bm{\theta}}) &= 1 - 
    \mathbb{E}_{\bm{x}\sim \mathcal{\bm{D}}_{test}}  \Big[ \mathds{1} (\max\{f_{\bm{\theta}}(\bm{x})\} \geq \tau)\Big] \\
    &= 1 - \mathbb{E}_{\bm{x}\sim \mathcal{\bm{D}}_{test}}  \Big[ \mathds{1} (\max \{\bm{\hat{p}(y)}\} \geq \tau)\Big] \\
    &= 1 - \mathbb{E}_{\bm{x}\sim \mathcal{\bm{D}}_{test}}  \Big[ \mathds{1} (\max \{\hat{p}(y_1),., \hat{p}(y_C)\} \geq \tau)\Big],
\end{split}
\end{equation}
where $\tau$ is the pre-defined confidence threshold, and a prediction will be rejected by the model if the predictive confidence over all categories could not achieve the threshold.
\end{definition}
Entropy $\mathcal{H}(f_{\bm{\theta}})$ measures the averaged uncertainty of a random variable's possible outcomes. Therefore, an ideal uncertainty estimation model will show a high entropy over all test (out-domain) data, whereas our attacker will increase the model's confidence for the wrong predictions, resulting in a reduced entropy. Rejection rate $\mathcal{R}(f_{\bm{\theta}})$ measures the fraction of the rejected test samples over the test set. Given the test set and a threshold $\tau$, a reasonable uncertain model should show a high rejection rate as it is too uncertain to make any predictions for the out-domain data.

Note that in Equation~\ref{eq:entropy} and Equation~\ref{eq:rejection_rate}, it is required that all $\hat{p}(y_i) \in \bm{\hat{p}(y)}$ are probabilities. Therefore, for the uncertain models which do not output a probability distribution, their outputs will be normalized into probabilities for evaluation in our experiments. For instance, since DUE outputs the RBF-based distances, we normalize all distances into a categorical distribution (More details in Section~\ref{sec:exp}), but Deep Ensemble uses averaged predicted probabilities directly \cite{lakshminarayanan2016simple}, meaning that there is no need for normalization for Deep Ensemble.

\subsection{Problem Formulation}
\paragraph{Threat Model.} Our threat model is that the adversarial attacker tries to perturb any out-domain test image $\bm{x} \sim \mathcal{P}_{\mathrm{OUT}}$ into $\bm{x}'$, so that the uncertainty estimation model will believe that $\bm{x}'$ is in-domain data and produce a high-confident prediction for it. We assume a white-box attack \cite{goodfellow2014explaining,galil2021disrupting}.
Moreover, we require that $\bm{x}'$ must be visually innocuous, indicating that $\bm{x}'$ and $\bm{x}$ must be geometrically close to each other.

To attack the uncertainty estimation on a specific out-domain test sample $\bm{x}$, we need to solve:
\begin{equation}
\label{eq:problem_optimization}
\begin{split}
    & \mathrm{min}_{\bm{\delta}} \Big[ \arg\max\{\bm{f}_{\bm{\theta}}(\bm{x} + \bm{\delta})\} \geq \tau \Big] \\
     & \mathrm{s.t.} \qquad \bm{x}\sim \mathcal{P}_{\mathrm{OUT}}, \qquad \|\bm{\delta}\| \leq \epsilon.
\end{split}
\end{equation}
In Equation~\ref{eq:problem_optimization}, $\tau$ represents the confidence threshold defined for Equation~\ref{eq:rejection_rate}. $\epsilon$ is a small non-negative scalar, bounding the norm of the adversarial perturbation, so that the perturbed image is visually innocuous. Under this formulation, the optimal adversarial perturbation could fool the uncertain model to make predictions with high-confidence (higher than the rejection threshold) without being visually detectable.

\section{Algorithm}
\label{sec:alg}
To successfully perturb out-domain test samples, our objective function (Equation~\ref{eq:problem_optimization}) must be solved rationally. However, due to the non-convexity of $\bm{f}_{\bm{\theta}}$ and non-differentiabiliy of $\arg \max$, it is intractable to compute a closed-form solution for Equation~\ref{eq:problem_optimization}. Therefore, instead of solving Equation~\ref{eq:problem_optimization} analytically, we propose to approximate optimal out-domain adversarial examples using a gradient-based method.

To begin with, we replace the optimization objective of Equation~\ref{eq:problem_optimization} with a new differentiable objective. That is, for an out-doman image $\bm{x}$, the attacker approximates its optimal perturbation by solving:
\begin{equation}
\label{eq:problem_optimization_new}
\begin{split}
    & \mathrm{min}_{\bm{\delta}} \Big[ l(\bm{f}_{\bm{\theta}}(\bm{x} + \bm{\delta}), \hat{y}) \Big] \\
     & \mathrm{s.t.} \qquad \bm{x}\sim \mathcal{P}_{\mathrm{OUT}}, \quad \|\bm{\delta}\| \leq \epsilon,
\end{split}
\end{equation}
where $\hat{y}$ is the heuristically found closest attacked label. More precisely, $\hat{y}$ is the original prediction produced by the uncertainty estimation model for the clean out-domain image.
\begin{equation}
    \begin{split}
    \label{eq:attacked_label}
     \hat{y} &= \arg \max_{y} \hat{\bm{p}}(y) \\
     & = \arg \max_{i} \Big[ \hat{p}(y_1),..., \hat{p}(y_i),...,\hat{p}(y_C) \Big] \\
      & = \arg \max_{i} \Big[ [f_{\bm{\theta}}(\bm{x})]_1,..., [f_{\bm{\theta}}(\bm{x})]_i,...,[f_{\bm{\theta}}(\bm{x})]_C \Big].
    \end{split}
\end{equation}
Note that it is determined by Equation~\ref{eq:attacked_label} that our proposed attack is untargeted. That being said, the adversary only moves the out-domain data to its closest data manifold instead of targeted ones. This untargeted formulation could guarantee the efficiency of the adversary: the path, over which the clean data is moved, could be much shorter compared to targeted attacks, since the targeted data manifolds could be further.

\begin{algorithm}[h]
\caption{Perturbing out-domain data}
\label{alg:algorithm}
\textbf{Input}: uncertainty estimation model $f_{\bm{\theta}}$, a clean test sample $\bm{x} \sim \mathcal{D}_{\mathrm{test}}$ \\
\textbf{Parameter}: adversarial radius $\epsilon$, number of iterations for the attack $K$, step size $\eta$ at each iteration \\
\textbf{Output}: the adversarial test sample $\bm{x}'$

\begin{algorithmic}[1] 
\STATE Initialize the iteration counter $k=0$
\STATE Initialize the adversarial example $\bm{x}'=\bm{x}$
\WHILE{$k \leq K$}
    \STATE $\hat{y} = \arg \max_{y} \hat{\bm{p}}(y)$
    \STATE $l(f_{\bm{\theta}}, \bm{x}', \hat{y}) = l(f_{\bm{\theta}}(\bm{x}'), \hat{y})$
    \STATE $\bm{x'} = \prod_{\mathbb{B}(\bm{x}, \epsilon)}(\bm{x}' - \eta \textsf{sign} \nabla_{\bm{x}'}l(f_{\bm{\theta}}, \bm{x}', \hat{y}))$, where $\prod$ is the projection operator \;
\ENDWHILE
\STATE \textbf{return} $\bm{x}'$
\end{algorithmic}
\end{algorithm}

The concrete procedure of generating an adversarial example for a out-domain sample $\bm{x}$ is presented in Algorithm~\ref{alg:algorithm}. In Algorithm~\ref{alg:algorithm}, Line 3 to Line 7 demonstrate how the optimal adversarial example (i.e. the minimizor of Equation~\ref{eq:problem_optimization_new}) will be approximated using projected gradient descent. In particular, Line 4 corresponds to Equation~\ref{eq:attacked_label}, namely the finding of the attacked label $\hat{y}$. Line 5 to Line 6 demonstrate how the adversarial example is updated using gradient of the loss evaluated at current steps. The projection operation $\prod$ guarantees that the norm of the adversarial perturbation would not exceed the adversarial radius $\epsilon$. After applying Algorithm~\ref{alg:algorithm} to an out-domain sample $\bm{x}$, the perturbed sample $\bm{x}'$ will be predicted by the uncertain model $f_{\bm{\theta}}$ as $\hat{y}$ with a high confidence.

Finally, we would like to comment on the differentiability of our formulation for attacking out-domain uncertainty estimation. Firstly, recall the reason we introduce Equation~\ref{eq:problem_optimization_new} to replace Equation~\ref{eq:problem_optimization} is that the operation $\arg \max$ in Equation~\ref{eq:problem_optimization} is not differentiable. Using a differentiable surrogate loss function $l$ (e.g. cross-entropy, variational ELBO), the gradient could be computed for the output layer of the uncertain model at the first step. However, different uncertainty estimation algorithms could be non-differentiable by design. For instance, in Gaussian process based algorithms (e.g. DUE \cite{van2021feature}), Monte Carlo sampling and other non-differentiable computation are usually employed to compute the final predictive uncertainty distribution. Therefore, to address such non-differentiable issues, we follow the exact implementation in the papers of DUE to study the feasibility of our proposed attack. For DUE \cite{van2021feature}, the reparametrization trick is implemented as in \cite{van2021feature} to make the generation of the adversarial perturbation differentiable. As for the Gaussian process in SNGP \cite{liu2020simple}, we follow the training implementation presented in \cite{liu2020simple} to guarantee the differentiability of the uncertain model. Regarding Deep Ensemble and DUQ, both of them are deterministic and differentiable, hence our attack could be launched directly.

\section{Experiments}
\label{sec:exp}
We evaluate the efficacy of our proposed out-domain uncertainty attack by assessing to which extent, the victim model could be deceived to make high-confident predictions for perturbed out-domain data. From a technical point of view, the victim models are trained to learn the in-domain data distribution using the in-domain training set $\mathcal{D}_{train}$ (as defined in Section~\ref{sec:pro}), but are tested on perturbed out-domain data.

\subsection{Dataset and Experimental Setup}
\paragraph{Dataset.} Following the experimental design presented in \cite{lakshminarayanan2016simple,van2020uncertainty,van2021feature} we use MNIST \cite{lecun1998gradient} vs. NotMNIST \footnote{dataset link: http://yaroslavvb.blogspot.co.uk/2011/09/notmnist-dataset.html}, and CIFAR-10 \cite{krizhevsky2009learning} vs.SVHN \cite{netzer2011reading} to test the efficacy of our proposed attack on corrupting the out-domain uncertainty estimation algorithms in DNNs. MNIST dataset contains 0-9 handwritten digit images, and NotMNIST dataset contains images for letters A-J (10 classes) taken from different fonts. CIFAR10 is a dataset containing images of common objects of 10 kinds in the real-world (airplane, automobile, dogs, e.t.c.), whereas SVHN consists of colored digit images taken from real-world scenes. Based on the description above, it is clear that MNIST vs. NotMNIST and CIFAR10 vs. SVHN represent drastically different data domains.

\paragraph{Experimental Setup.} \footnote{Code will be released after acceptance of this paper.} In many out-domain uncertainty estimation studies \cite{lakshminarayanan2016simple,van2020uncertainty,van2021feature}, it is common to train uncertainty estimation models with in-domain data, and then evaluate the predictive confidence returned by the model over the out-domain test data. Therefore, without loss of generality, we also follow the experimental setup for out-domain uncertainty estimation in \cite{lakshminarayanan2016simple,van2020uncertainty,van2021feature}. There are two sets of experiments in our work. In the first set of experiment, the uncertain models are trained using MNIST (in-domain), and tested on NotMNIST (out-domain). In the second set of experiments, CIFAR10 is used as the in-domain training data, and the resulted model is tested on out-domian SVHN data.

\subsection{Baselines}
\paragraph{Baseline DNN Architectures.} To efficiently model the data distribution and avoid overfitting, we select one commonly used neural network architecture for two sets of experiments respectively. In particular, for MNIST vs. NotMNIST, LeNet-5 \cite{lecun1998gradient} is used as the baseline architecture to build uncertainty estimation models. As for CIFAR10 vs. SVHN, we use ResNet-18 \cite{he2016deep}. Note that the baseline architectures are not equivalent to the final uncertainty estimation models. For instance, in a Deep Ensemble model, there are multiple ensemble classifiers. For DUQ, DUE and SNGP, the baseline models only work as feature extractors, after which there could still be further computational modules, such as RBF kernels or Gaussian process.

\paragraph{Baseline Uncertainty Estimation Algorithms.} To show the feasibility and efficacy of our proposed attack algorithm, 4 uncertainty estimation algorithms are included in our experiments, namely Deep Ensemble \cite{lakshminarayanan2016simple}, DUQ \cite{van2020uncertainty}, DUE \cite{van2021feature} and SNGP \cite{liu2020simple}. 
For all experiments, the key hyperparameters for our method and all baseline algorithms are tuned to achieve their best performance for a fair comparison. In summary, we have following trained models under different training schemes:
\begin{itemize}
    \item \textbf{Deep Ensemble:} The vanilla deep ensemble for uncertainty estimation \cite{lakshminarayanan2016simple}. 
    \item \textbf{Deep Ensemble Adv.:} The augmented version of deep ensemble using adversarially trained ensemble networks \cite{lakshminarayanan2016simple}.
    \item \textbf{DUQ:} Uncertainty estimation using RBF kernels \cite{van2020uncertainty}.
    \item \textbf{DUE:} Uncertainty estimation using Gaussian process and variational ELBO \cite{van2021feature}.
    \item \textbf{SNGP:} Spectral-normalized Gaussian process for uncertainty estimation \cite{liu2020simple}.
\end{itemize}

Note that there exists another augmented variant of Deep Ensemble. That is, the augmented deep ensemble model could be built upon a set of adversarially trained ensemble networks. Following the configuration in the original paper of Deep Ensemble \cite{lakshminarayanan2016simple}, we train the ensemble networks with fast gradient sign method (FGSM) instead of projected gradient descent (PGD). For a fair comparison, when evaluating deep ensemble models, we also modify our attack with iterations to a single-step attack. When performing adversarial training, the training radius $\epsilon_{tr}$ must be specified. In our experiments, two different training adversarial radius ($\epsilon_{tr}$) are used. For MNIST vs. NotMNIST, adversarial training with $\epsilon_{tr}=0.1$ and $\epsilon_{tr}=0.2$ are performed, whereas for CIFAR10 vs. SVHN, the training adversarial radius $\epsilon_{tr}$ is set to $0.016$ and $0.031$, respectively.

\subsection{Evaluation Results}
\paragraph{Evaluating Efficacy.} To begin with, we conduct two sets of experiments to evaluate the efficacy of our proposed adversarial attack. The numerical results on entropy and rejection rate are reported in Table~\ref{tab:core_comparison}. $\mathcal{H}_{clean}$ and $\mathcal{R}_{clean}$ are the entropy and rejection rate computed using the clean, unperturbed out-domain data. In contrast, $\mathcal{H}_{adv}$ and $\mathcal{R}_{adv}$ are evaluated using perturbed out-domain data. In both MNIST vs. NotMNIST and CIFAR10 vs. SVHN, the confidence threshold $\tau$ for computing $\mathcal{R}_{clean}$ and $\mathcal{R}_{adv}$ is $0.9$. That is, the uncertain model only makes predictions when the confidence score is greater than $0.9$. Moreover, to attack NotMNIST images, the adversarial radius $\epsilon$ is set to $0.1$, and $\epsilon=0.016$ for SVHN images.

\begin{table}[th]
    \centering
    \begin{subtable}[h]{\columnwidth}
        \centering
        \resizebox{0.9\columnwidth}{!}{
        \begin{tabular}{c|cc|cc}
        \hline
        \toprule
        & \hfil $\mathcal{H}_{clean}$ & \hfil $\mathcal{H}_{adv}$ & \hfil $\mathcal{R}_{clean}$  & \hfil $\mathcal{R}_{adv}$  \\
        \hline
        \hline
        \hfil D.E. & \hfil 0.42 & \hfil \textbf{0.26} & \hfil 0.44 & \hfil \textbf{0.26} \\
        \hfil D.E.Adv. ($\epsilon_{tr}=0.1$) & \hfil 0.65 & \hfil \textbf{0.49} & \hfil 0.60 & \hfil \textbf{0.44} \\
        \hfil D.E.Adv. ($\epsilon_{tr}=0.2$) & \hfil 0.72 & \hfil \textbf{0.58} & \hfil 0.65 & \hfil \textbf{0.52} \\
        \hfil DUQ & \hfil 1.29 & \hfil \textbf{1.02} & \hfil 0.98 & \hfil \textbf{0.78} \\
        \hfil DUE & \hfil 0.83 & \hfil \textbf{0.09} & \hfil 0.69 & \hfil \textbf{0.01} \\
        \hfil SNGP. & \hfil 1.22 & \hfil \textbf{0.23} & \hfil 0.80 & \hfil \textbf{0.17} \\
        \hline
        \toprule
        \end{tabular}
        }
        \caption{MNIST vs. NotMNIST: 10-iteration adversarial attack with $\epsilon=0.1$, rejection rate computed with confidence threshold $\tau=0.9$.}
        \label{tab:mnist_core}
    \end{subtable}
    
    \begin{subtable}[h]{\columnwidth}
        \centering
        \resizebox{0.95\columnwidth}{!}{
        \begin{tabular}{c|cc|cc}
        \hline
        \toprule
        & \hfil $\mathcal{H}_{clean}$ & \hfil $\mathcal{H}_{adv}$ & \hfil $\mathcal{R}_{clean}$ & \hfil $\mathcal{R}_{adv}$  \\
        \hline
        \hline
        \hfil D.E. & \hfil 0.45 & \hfil \textbf{0.34} & \hfil 0.53 & \hfil \textbf{0.43} \\
        \hfil D.E.Adv. ($\epsilon_{tr}=0.016$) & \hfil 1.01 & \hfil \textbf{0.65} & \hfil 0.93 & \hfil \textbf{0.65} \\
        \hfil D.E.Adv. ($\epsilon_{tr}=0.032$) & \hfil 1.30 & \hfil \textbf{1.03} & \hfil 0.98 & \hfil \textbf{0.90} \\
        \hfil DUQ & \hfil 1.32 & \hfil \textbf{0.56} & \hfil 1.00 & \hfil \textbf{0.72} \\
        \hfil DUE & \hfil 1.26 & \hfil \textbf{1.05} & \hfil 0.95 & \hfil \textbf{0.81} \\
        \hfil SNGP. & \hfil 1.04 & \hfil \textbf{0.04} & \hfil 0.90 & \hfil \textbf{0.02} \\
        \hline
        \toprule
        \end{tabular}
        }
        \caption{CIFAR10 vs. SVHN: 10-iteration adversarial attack with $\epsilon=0.016$, rejection rate computed with confidence threshold $\tau=0.9$.}
        \label{tab:cifar_core}
    \end{subtable}
    \caption{Evaluating the efficacy of proposed attack algorithm on MNIST vs. NotMNIST and CIFAR10 vs. SVHN.}
    \label{tab:core_comparison}
    \vspace{-0.4cm}
\end{table}

From Table~\ref{tab:core_comparison}, we observe that our proposed attack could reduce both the entropy and rejection rate of different uncertain models significantly. This indicates that the uncertain models would be fooled by our attack to be over-confident about the wrong predictions they made for the out-domain data. For instance, regarding the first evaluation metric, the entropy on NotMNIST of SNGP is reduced from 1.22 to 0.23. As for the second evaluation metric, when making predictions for clean NotMNIST images, the DUQ model would reject 98\% of the predictions due to the lack of confidence. However, when attacked by our algorithm, the rejection rate of DUQ would be reduced to 78\%. Similar observations are made on all other baseline uncertain models in terms of both entropy and rejection rate. In other words, the uncertain models would be less uncertain and could fail to reject many predictions, which originally would have been rejected when there is no attack. As for CIFAR10 vs. SVHN, our attack still shows high efficacy in terms of increasing the uncertain models' confidence on the out-domain data according to Table~\ref{tab:cifar_core}. For example, the DUQ model could achieve the rejection rate of 100\% on clean SVHN images, but fails to reject 28\% perturbed images. Moreover, when comparing Table~\ref{tab:mnist_core} and Table~\ref{tab:cifar_core}, we also notice that different uncertainty estimation algorithms show different robustness on different datasets. For instance, the uncertainty of DUE could be corrupted dramatically on MNIST vs. NotMNIST (the entropy is reduced from 0.83 to 0.09), but on CIFAR10 vs. SVHN, DUE shows a much stronger robustness (the entropy is reduced merely from 1.26 to 1.05). In contrast, DUQ is much more robust against our attack on MNIST vs. NotMNIST, but more vulnerable on CIFAR10 vs. SVHN.

To summarize, the consistent reduction of entropy and rejection rates on all uncertain models in Table~\ref{tab:core_comparison} verifies 
that the state-of-the-art uncertainty estimation algorithms could be extremely vulnerable in the presence of adversarial attacks. Moreover, despite the vulnerability of all tested uncertainty algorithms, they also display different robustness against the attacks on different datasets.

\begin{table}[th]
    \centering
    \begin{subtable}[h]{\columnwidth}
        \centering
        \resizebox{\columnwidth}{!}{
        \begin{tabular}{c|c|c|c|c|c|c}
        \hline
        \toprule
        & \hfil D.E. & \hfil D.E. Adv. & \hfil D.E. Adv. & \hfil DUQ & \hfil DUE  & \hfil SNGP \\
        & \hfil  & \hfil $\epsilon_{tr}=0.1$ & \hfil $\epsilon_{tr}=0.2$ &  &  &  \\
        \hline
        \hline
        $\mathcal{H}_{clean}$ & \hfil 0.42 & \hfil 0.65 & \hfil 0.72 & \hfil 1.29 & \hfil 0.83 & \hfil 1.22   \\
        \hline
        $\mathcal{H}_{adv}$  & \hfil \textbf{0.26} & \hfil \textbf{0.49} & \hfil \textbf{0.58} & \hfil \textbf{1.02} & \hfil \textbf{0.09} & \hfil \textbf{0.23} \\
        $\mathcal{H}'_{adv}$  & \hfil \textbf{0.22} & \hfil \textbf{0.45} & \hfil \textbf{0.53} & \hfil \textbf{0.80} & \hfil \textbf{0.07} & \hfil \textbf{0.14} \\
        $\mathcal{H}''_{adv}$  & \hfil \textbf{0.22} & \hfil \textbf{0.48} & \hfil 0.73 & \hfil \textbf{0.72} & \hfil \textbf{0.08} & \hfil \textbf{0.15} \\
        \hline
        \hline
        $\mathcal{R}_{clean}$ & \hfil 0.44 & \hfil 0.60 & \hfil 0.65 & \hfil 0.98 & \hfil 0.69 & \hfil 0.80  \\
        \hline
        $\mathcal{R}_{adv}$ & \hfil \textbf{0.26} & \hfil \textbf{0.44} & \hfil \textbf{0.52} & \hfil \textbf{0.78} & \hfil \textbf{0.01} & \hfil \textbf{0.17} \\
        $\mathcal{R}'_{adv}$ & \hfil \textbf{0.22} & \hfil \textbf{0.40} & \hfil \textbf{0.47} & \hfil \textbf{0.48} & \hfil \textbf{0.00} & \hfil \textbf{0.12} \\
        $\mathcal{R}''_{adv}$ & \hfil \textbf{0.21} & \hfil \textbf{0.41} & \hfil \textbf{0.57} & \hfil \textbf{0.40} & \hfil \textbf{0.00} & \hfil \textbf{0.11} \\
        \toprule
        \end{tabular}
        }
        \caption{MNIST vs. NotMNIST: 10-iteration adversarial attack with $\epsilon=0.1$, $\epsilon=0.2$ and $\epsilon=0.3$, rejection rate computed with confidence threshold $\tau=0.9$.}
        \label{tab:mnist_robust}
    \end{subtable}
    
    \begin{subtable}[h]{\columnwidth}
        \centering
        \resizebox{\columnwidth}{!}{
        \begin{tabular}{c|c|c|c|c|c|c}
        \hline
        \toprule
        & \hfil D.E. & \hfil D.E. Adv. & \hfil D.E. Adv. & \hfil DUQ & \hfil DUE  & \hfil SNGP \\
        & \hfil  & \hfil $\epsilon_{tr}=0.016$ & \hfil $\epsilon_{tr}=0.031$ &  &  &  \\
        \hline
        \hline
        $\mathcal{H}_{clean}$ & \hfil 0.45 & \hfil 1.01 & \hfil 1.30 & \hfil 1.32 & \hfil 1.26 & \hfil 1.04   \\
        \hline
        $\mathcal{H}_{adv}$ & \hfil \textbf{0.34} & \hfil \textbf{0.65} & \hfil \textbf{1.03} & \hfil \textbf{0.56} & \hfil \textbf{1.05} & \hfil \textbf{0.04} \\
        $\mathcal{H}'_{adv}$ & \hfil {0.51} & \hfil \textbf{0.49} & \hfil \textbf{0.86} & \hfil \textbf{0.32} & \hfil \textbf{0.98} & \hfil \textbf{0.03} \\
        $\mathcal{H}''_{adv}$ & \hfil {0.68} & \hfil \textbf{0.44} & \hfil \textbf{0.75} & \hfil \textbf{0.24} & \hfil \textbf{0.93} & \hfil \textbf{0.03} \\
        \hline
        \hline
        $\mathcal{R}_{clean}$ & \hfil 0.53 & \hfil 0.93 & \hfil 0.98 & \hfil 1.00 & \hfil 0.95 & \hfil 0.90 \\
        \hline
        $\mathcal{R}_{adv}$ & \hfil \textbf{0.39} & \hfil \textbf{0.65} & \hfil \textbf{0.90} & \hfil \textbf{0.72} & \hfil \textbf{0.81} & \hfil \textbf{0.02} \\
        $\mathcal{R}'_{adv}$ & \hfil {0.59} & \hfil \textbf{0.48} & \hfil \textbf{0.80} & \hfil \textbf{0.13} & \hfil \textbf{0.80} & \hfil \textbf{0.02} \\
        $\mathcal{R}''_{adv}$ & \hfil {0.76} & \hfil \textbf{0.42} & \hfil \textbf{0.70} & \hfil \textbf{0.04} & \hfil \textbf{0.82} & \hfil \textbf{0.00} \\
        \toprule
        \end{tabular}
        }
        \caption{CIFAR10 vs. SVHN: 10-iteration adversarial attack with $\epsilon=0.016$,$\epsilon=0.031$,$\epsilon=0.063$ rejection rate computed with confidence threshold $\tau=0.9$.}
        \label{tab:cifar_robust}
    \end{subtable}
    \caption{Evaluating the efficacy of proposed attack algorithm on MNIST vs. NotMNIST and CIFAR10 vs. SVHN.}
    \label{tab:robust_comparison}
    \vspace{-0.3cm}
\end{table}

\paragraph{Robustness Study.} In addition to the efficacy evaluation of our proposed attack shown in Table~\ref{tab:cifar_core}, we also conduct a set of robustness study to understand the relationship between the adversarial radius and attack effect. In particular, we changed the adversarial radius $\epsilon$  from $0.1$ (corresponding to $\mathcal{H}_{adv}$ and $\mathcal{R}_{adv}$) to $0.2$ (corresponding to $\mathcal{H}'_{adv}$ and $\mathcal{R}'_{adv}$) and $0.3$ (corresponding to $\mathcal{H}''_{adv}$ and $\mathcal{R}''_{adv}$) for NotMNIST. Similarly, for SVHN, $\epsilon$ is changed from $0.016$ to $0.031$ and $0.063$. As shown in Table~\ref{tab:mnist_robust} and in Table~\ref{tab:cifar_robust}, we observe that in general, with larger $\epsilon$, the attack becomes more powerful and the uncertainty of the model would be further reduced (bold numbers). For instance, the entropy of DUQ on NotMNIST is consistently reduced from $1.02$ to $0.80$ and $0.72$ with the increase of $\epsilon$. However, an opposite trend is also observed, such as on SVHN, both entropy and rejection rate of the vanilla deep ensemble increase when $\epsilon$ becomes larger. We believe the optimal $\epsilon$ for attacking different dataset is at different ranges.

\begin{figure}[ht]
\begin{subfigure}{\columnwidth}
  \centering
  \includegraphics[width=\linewidth]{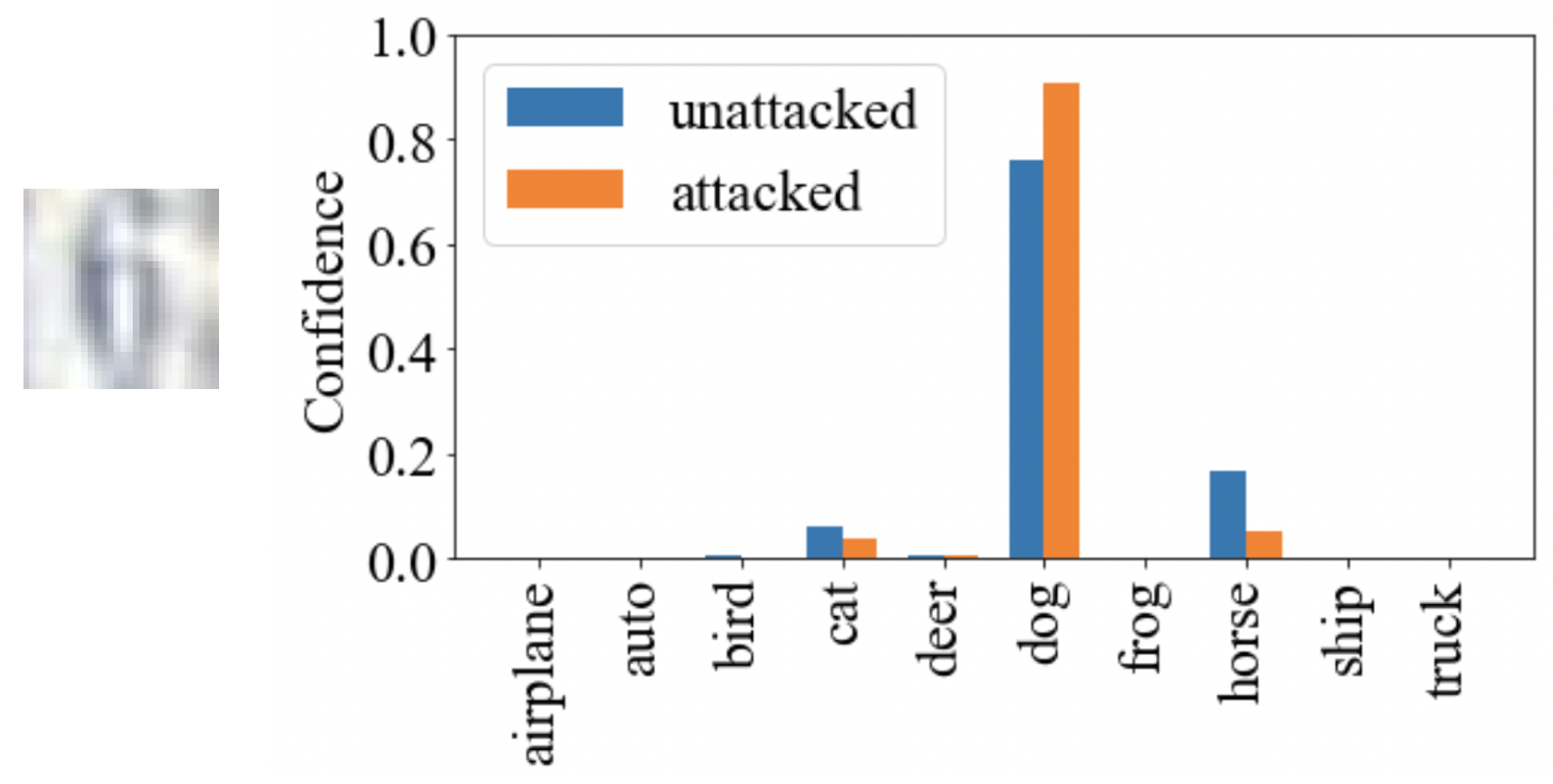}
  \caption{Attack adversarially trained Deep Ensemble (D.E.Adv. $\epsilon_{tr}=0.016$) with $\epsilon=0.016$. The clean image of the digit '6' from SVHN dataset will be recognized by the model as a dog. After our attack, the confidence is higher than 90\%.}
  \label{fig:confidence_visualization_1}
\end{subfigure}
\begin{subfigure}{\columnwidth}
  \centering
  \includegraphics[width=\linewidth]{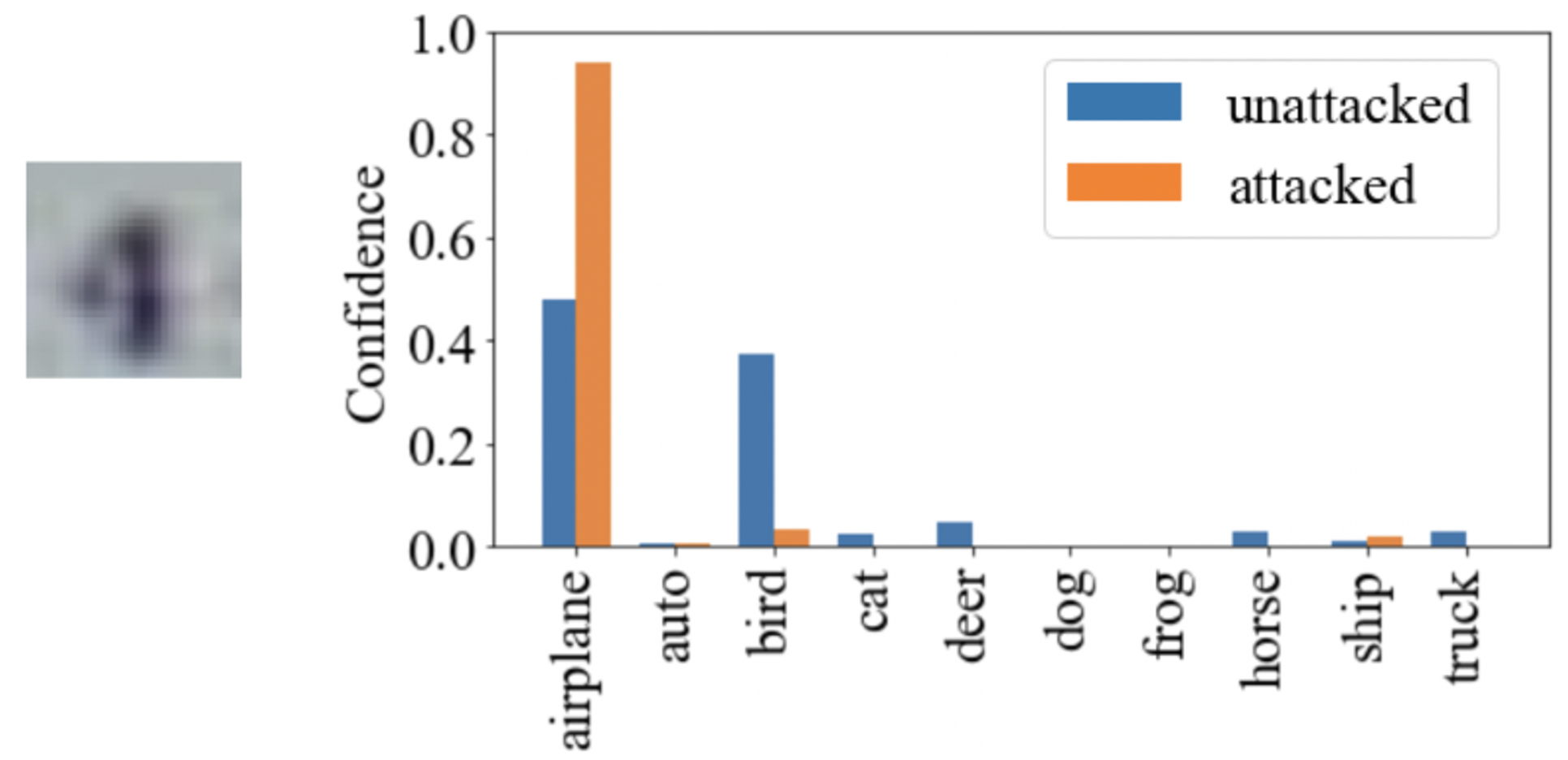}
  \caption{Attack DUE with $\epsilon=0.031$. The clean image of the digit '4' from SVHN dataset will be recognized by the model as an airplane. After our attack, the confidence is higher than 90\%.}
  \label{fig:confidence_visualization_2}
\end{subfigure}
\caption{The predictive confidence distribution over the in-domain categories.}
\label{fig:confidence_visualization}
\vspace{-0.3cm}
\end{figure}

\paragraph{Visualization of Confidence Distribution.} Due to the limited space, we randomly pick two of attacked SVHN images and show the shift of the confidence distribution after applying our attack. 
In Figure~\ref{fig:confidence_visualization_1}, an adversarially train deep ensemble model is attacked using our proposed algorithm, and in Figure~\ref{fig:confidence_visualization_2}, a DUE model is attacked. Note that both uncertain models are trained using CIFAR10 as in-domain data, and are tested on SVHN images. As we can observe, both uncertain models are fooled by the adversarial examples to make high-confidence wrong predictions for the out-domain data. In Figure~\ref{fig:confidence_visualization_2}, the DUE model is only around 50\% confident about the prediction when there is no attack. However, when the image is perturbed, the confidence is raised to higher than 90\%. Originally, the uncertain prediction on the unperturbed images could have been rejected by model, and further deferred the image to an expert, but under our attack, the model will be extremely confident about the wrong predictions, and no further critical analysis would be performed.

\section{Conclusion}
\label{sec:conclu}
With the intention of raising the attention on the topic of the robustness of uncertainty estimation, we investigated and designed a prototype white-box adversarial attacks on \emph{out-domain} uncertainty estimation. As shown in our experiments, state-of-the-art uncertainty estimation algorithms could be deceived easily by our proposed adversary to make predictions for the out-domain images with very high confidence. In terms of the applications with real-world consequences, the vulnerability of the uncertain models could open a fatal loophole for attackers, leading to catastrophic consequences. Therefore, despite the accuracy of the uncertainty estimation, it is also important to improve the robustness of various uncertainty estimation algorithms in DNNs.

\section*{Ethics Statement}
\paragraph{Adversarial attacks could cause extremely high threat to existing uncertainty estimation in deep neural networks.}
Despite the impressive performance on uncertainty estimation on out-domain data, state-of-the-art algorithms still fail catastrophically under the attack of imperceptible perturbations. 
The existence of such ``out-domain'' adversarial examples exposes a serious vulnerability in current uncertainty-based ML systems, such as autonomous driving, medical diagnosis and digital financial systems. In these applications with real-world consequences, the vulnerability of uncertainty systems could place our lives and security at risk. 

\paragraph{Our work has the potential to inspire future studies on a new type of adversarial defense mechanism and the design of robust uncertainty estimation algorithms.}
Although in this work, we present a simple prototype adversarial attack on out-domain uncertainty estimation, the core intention of this work is to raise the attention of the research community to systematically investigate the robustness of the uncertainty estimation in DNNs. We believe our attack could be used to test future robust uncertainty estimation algorithms.

\section*{Acknowledgments}

This research is supported in part by the National Science Foundation under Grant No. CHE-2105032,  IIS-2008228, CNS-1845639, CNS-1831669. The views and conclusions contained in this document are those of the authors and should not be interpreted as representing the official policies, either expressed or implied, of the U.S. Government. The U.S. Government is authorized to reproduce and distribute reprints for Government purposes notwithstanding any copyright notation here on.

\bibliographystyle{named}
\bibliography{reference}

\end{document}